\documentclass[11pt,a4paper]{article}
\usepackage[hyperref]{emnlp2020}
\usepackage{times}
\usepackage{latexsym}
\usepackage{tabularx}
\usepackage{booktabs}
\usepackage{graphicx, amssymb, amsmath, amsfonts, color, array, ulem, xspace}
\usepackage{siunitx}
\usepackage{CJKutf8}
\usepackage{dblfloatfix}

\usepackage{microtype}

\aclfinalcopy 

\DeclareMathOperator{\E}{\mathbb{E}}

\DeclareMathOperator*{\argmax}{argmax}
\DeclareMathOperator*{\softmax}{softmax}

\newcommand \ndomain {K}
\newcommand \srcenc {E_{\mathrm{src}}}
\newcommand \tgtenc {E_{\mathrm{tgt}}}
\newcommand \dec {D}
\newcommand \BLEU{\text{BLEU}}

\newcommand \pairwise{\texttt{pairwise}\xspace}
\newcommand \mbleu{\texttt{mBLEU}\xspace}

\newcommand*{\bleu}[1]{\num[round-mode=places,round-precision=1]{#1}}

\title{Target Conditioning for One-to-Many Generation}

\author{Marie-Anne Lachaux \\
 Facebook AI Research \\
  \texttt{malachaux@fb.com} \\
  \And 
  Armand Joulin \\
  Facebook AI Research \\
  \texttt{ajoulin@fb.com} \\
  \And 
  Guillaume Lample\\
  Facebook AI Research \\
  \texttt{glample@fb.com} \\}

\date{}

\begin{document}
\maketitle

\begin{abstract}
Neural Machine Translation (NMT) models often lack diversity in their generated translations, even when paired with search algorithm, like beam search.
A challenge is that the diversity in translations are caused by the variability in the target language, and cannot be inferred from the source sentence alone.
In this paper, we propose to explicitly model this one-to-many mapping by conditioning the decoder of a NMT model on a latent variable that represents the domain of target sentences.
The domain is a discrete variable generated by a target encoder that is jointly trained with the NMT model.
The predicted domain of target sentences are given as input to the decoder during training.
At inference, we can generate diverse translations by decoding with different domains.
Unlike our strongest baseline \cite{shen2019mixture}, our method can scale to any number of domains without affecting the performance or the training time.
We assess the quality and diversity of translations generated by our model with several metrics, on three different datasets.
\end{abstract}

\newcommand{\insertgenerationscomparison}{
\begin{table*}[b]
\centering
\resizebox{\textwidth}{!}{
\begin{tabular}{lll}
\toprule
    Source                 & \begin{CJK}{UTF8}{gbsn}参与投票的成员中,58\%反对该合同交易。\end{CJK}        & \begin{CJK}{UTF8}{gbsn}自11月份开始，俄罗斯民意也有所扭转。\end{CJK}        \\
    Human references       & It was rejected by 58\% of its members who voted in the ballot.          & Russian public opinion has also turned since November.                  \\
                           & Of the members who voted, 58\% opposed the contract transaction.         & Russian public opinion has started to change since November.            \\
                           & Of the members who participated in the vote, 58\% opposed the contract.  & The polls in Russian show a twist turn since the beginning of November. \\
    \\
    Beam 3, Top 3          & Of those voting, 58 per cent opposed the contract deal.                  & Since November, Russian public opinion has also turned around. \\
                           & Fifty-eight per cent of the members voting opposed the contract deal.    & Since November, Russian public opinion has also changed.       \\
                           & Fifty-eight per cent of the members voting opposed the contract.         & Russian public opinion has also changed since November.        \\
    \\
    Mixture of Experts     & Of the members who voted, 58\% opposed the deal.                         & Since November, the mood in Russia has also reversed. \\
    \cite{shen2019mixture} & Fifty-eight per cent of the members who voted opposed the contract deal. & Since November, opinion in Russia has also reversed.  \\
                           & Fifty-eight per cent of the voting members opposed the contract deal.    & Opinion in Russia has also shifted since November.    \\
    \\
    Our Model              & Of the members voting, 58 per cent opposed the contract deal.            & Since November, Russian public opinion has also reversed.                \\
                           & Fifty-eight per cent of the members who voted opposed the contract deal. & The mood in Russia has also reversed since November.                     \\
                           & Fifty-eight per cent of those voting had opposed this contract deal.     & There has also been a reversal in Russian public opinion since November. \\
\bottomrule
\end{tabular}
}
\caption{Two examples of generations by our model and different baselines on the WMT'17 Zh-En dataset. Beam search generation lack diversity. The target encoder model gives the most diverse sets of translations.}
\label{tab:generations_comparison}
\end{table*}
}

\newcommand{\insertgenerationszhen}{
\begin{table*}[h]
\centering
\resizebox{\textwidth}{!}{
\begin{tabular}{lll}
\toprule
Source    & \begin{CJK}{UTF8}{gbsn}完成该项作业需把北部高架引桥南移700毫米。\end{CJK}                               \\
Reference & This was done by pulling the northern approach viaduct 700 millimetres southwards.                      \\
Our model & Completion of the operation required the southern transfer of 700 mm from the northern elevated bridge. \\
          & This operation will require moving the northern elevated bridge to a further 700 mm south.              \\
          & The operation was completed by moving the northern elevated bridge to the south by 700 mm.              \\
\\
Source    & \begin{CJK}{UTF8}{gbsn}尽管该桥的规模和重量都很巨大，但是完成高架桥和桥面之间的合龙却需要精细作业，需要非常精准的装配。\end{CJK}                     \\
Reference & Although the bridge is very large and heavy, joining the viaduct and roadway is a meticulous operation, requiring precise assembly.                  \\
Our model & Despite the scale and weight of the bridge, the completion of the Alloy bridge and the deck requires fine work and very precise assembly.            \\
          & Although it is huge in size and weight, it requires fine work and very precise assembly to complete the Alloy between the viaduct and the deck.      \\
          & The bridge, though of great scale and weight, requires fine work and a very precise assembly to complete the Alloy between the viaduct and the deck. \\
\\
Source    & \begin{CJK}{UTF8}{gbsn}据报道，雷杰普·塔伊普·埃尔多安则五次提到“好朋友”普京。\end{CJK}        \\
Reference & According to the report, Recep Tayyip Erdoğan mentioned his ``good friend'' Putin five times. \\
Our model & It was reported that Recep Tayyip Erdoğan made five references to ``good friend'' Putin.      \\
          & Recep Tayyip Erdoğan reportedly mentioned ``good friend'' Putin five times.                   \\
          & The five-time mentions of ``good friend'' Putin were reported by Recep Tayyip Erdoğan.        \\
\\
Source    & \begin{CJK}{UTF8}{gbsn}叙利亚是关键因素之一。莫斯科近日在叙利亚扮演和事佬的角色，而俄罗斯与土耳其却支持相反派别。\end{CJK} \\
Reference & Syria is a key factor. Moscow has recently played the role of a mediator in Syria. However, Russia and Turkey supported opposing factions. \\
Our model & One of the key factors is Syria, where Moscow has played a conciliatory role, while Russia and Turkey have supported the opposite.         \\
          & Syria is one key factor, with Moscow playing a conciliatory role in Syria, while Russia and Turkey have supported the opposite.            \\
          & The role of Moscow as a peacemaker in Syria recently was one of the key factors, while Russia and Turkey supported the opposition.         \\
\\
Source    & \begin{CJK}{UTF8}{gbsn}双方就如何在叙利亚问题上求同存异未达成明确共识。\end{CJK}     \\
Reference & There is no clear consensus on where they can seek common ground on Syria.           \\
Our model & There is no clear consensus on how to seek common ground on Syria.                   \\
          & No clear consensus has been reached on how to find common ground on Syria.           \\
          & The two sides did not reach a clear consensus on how to seek common ground on Syria. \\
\\
Source    & \begin{CJK}{UTF8}{gbsn}莫斯科当时的反应很激烈\end{CJK} \\
Reference & Moscow's reaction was intense.      \\
Our model & Moscow was reacting violently.      \\
          & Moscow's reaction was very strong.  \\
          & The reaction in Moscow was intense. \\
\\
Source    & \begin{CJK}{UTF8}{gbsn}令人欣慰的是，一切都进展得很顺利。\end{CJK} \\
Reference & It is gratifying that everything is going well.         \\
Our model & It is gratifying to note that everything is going well. \\
          & Thankfully, everything has gone well.                   \\
          & To the relief of all, everything was going well.        \\
\bottomrule
\end{tabular}
}
\caption{Examples of generations by our model on the WMT'17 Zh-En dataset. Translations are sorted by domain index. The model generates high-quality translations with high diversity. Unlike beam search decoding, that tends to return similar hypotheses with only minor differences, our model generates various translations with very different prefixes, even for long sentences.}
\label{tab:generations_deen}
\end{table*}
}

\begin{figure*}[t!]
  \centering
  \includegraphics[width=0.85 \linewidth]{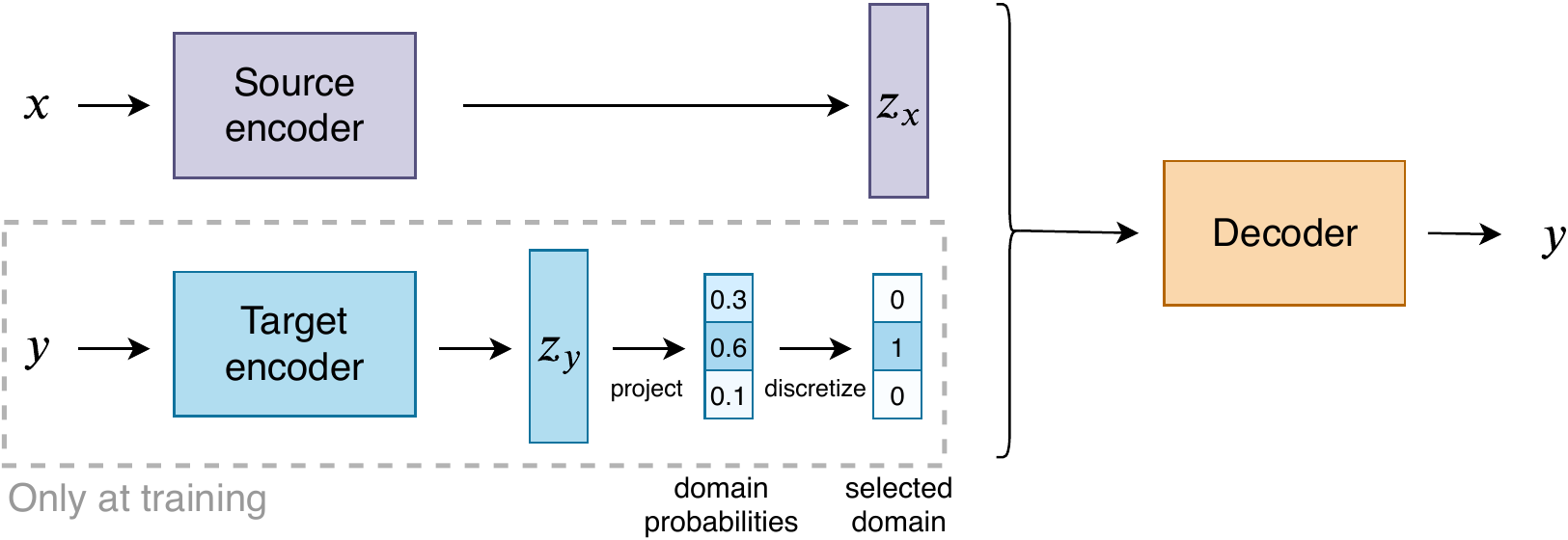}
  \caption{
  \textbf{Illustration of our model.} The model is composed of a source and a target encoder, and a decoder. At training time, a target sentence is encoded with the target transformer encoder to get a latent representation $z_y$.
The latent representation is linearly mapped to a vector of size $\ndomain$ on which apply a softmax to obtain domain probabilities.
Each domain is associated with an embedding.
The decoder is fed with both the source encoding, and the sum of the domain embeddings reweighted by their probabilities.
During inference, we can generate $\ndomain$ different hypotheses by switching the domain embedding that is fed to the decoder.
To prevent a train-test discrepancy, during training we apply an $\argmax$ operator on domain probabilities, with probability $p_{\textrm{hard}}$.
  }
  \label{figure:model}
\end{figure*}

\section{Introduction}

Neural Machine Translation (NMT) models are trained to translate a sentence from a source language into a target language.
There are many translations of the same sentence that are both grammatically correct and faithful to the source, but these translations may differ greatly in their vocabulary, style or grammar.
Inferring the best translation among them requires to explore a vast output space to cover this variability.
This is typically handle as a post-processing step using a search algorithm, like beam search.
This procedure is known to produce translations that lack in diversity, often differing only by a punctuation or a word~\cite{kumar2004minimum,li2016simple}.
While the search algorithm can certainly be improved, part of the problem resides also in the training of the NMT models; they are trained on 1-to-1 translation datasets without any objective to encourage diverse translations.

There are many ways to model the diversity of translations from data that contain only one translation, such as mixture of experts~\cite{shen2019mixture} or variational autoencoders~\cite{zhang2016variational}.
A particularity of machine translation is that it is a one-to-many mapping problem.
This means that the variability should be encoded by the target sentence and the question is how to combine a NMT system with a target sentence encoder with no posterior collapse.

In this work, we propose to combine the encoder of the NMT with a discrete target encoder.
Similar to other discrete autoencoders~\cite{kaiser2018fast,van2017neural}, each target sentence is assigned to a discrete variable, or domain, and each domain is associated with an embedding.
The embeddings from both encoders are then fed to the decoder of the NMT to form a translation.
The discrete latent representation follows a categorical distribution that is constrained to be uniform over the dataset to avoid a mode collapse. 
Since each domain has its own embedding, changing the domain embedding changes the translation.
At test time, we can thus condition the generation on each domain embedding to produce multiple translations with high diversity. 

Our approach is general and can be applied on top of any model with little computational overhead. 
An advantage of our approach is that the number of domains can be arbitrarily large without affecting the performance or the running time. 
Our approach can replace or work with beam search during inference.
We assess the quality and diversity of translations generated by our model with several metrics, on three different datasets.

\section{Related Work}

Several studies have proposed to sample diverse sequences by changing the value of a latent variable.
For example, one possibility is to add noise to the latent space of a Variational Auto-Encoder~\cite{kingma2013auto} to diversify samples in machine translation~\cite{zhang2016variational}, language modeling~\cite{bowman2015generating} or question generation~\cite{jain2017creativity}.
In particular,~\citet{zhang2016variational} also condition the decoder of a NMT Model on a target encoder.
As opposed to our work, the output of their encoder is continuous and sampling diverse generation requires to inject random noise, while we obtain diversity by switching between discrete domains. 
Similar noise injection mechanisms have been investigated to improve the diversity of responses in dialogue~\cite{serban2017hierarchical, cao2017latent, wen2017latent}, and image captioning~\cite{wang2017diverse, dai2017towards}.
Closer to our work, \cite{shen2019mixture,shu2019generating} and \citet{xu2018d} use domain embeddings to condition their generations. Unlike us, they do not condition the domain on the target, but select the domain which minimizes the reconstruction loss, which becomes expensive as the number of domains increases.
Another relevant work is the fast decoder of~\citet{kaiser2018fast} where they also combine a discrete encoder applied on the target sentence with the NMT encoder.
Their goal is to accelerate the decoding process of a machine translation system, while we are interested in efficiently sampling diverse translations. 


Another line of work focuses on improving the generation by changing the decoding scheme during inference~\cite{li2016simple, gu2017trainable} or by matching the training of the model to the decoding scheme~\cite{wiseman2016sequence,collobert2019fully}.
This is done by either training through a beam search decoder~\cite{wiseman2016sequence, collobert2019fully} or by reframing generation as a reinforcement learning problem~\cite{bengio2015scheduled,ranzato2015sequence}.
These works focus on the decoding scheme to improve generation, but do not address the problem of diversifying the outputs generated from the same input.

\section{Model}

In this section, we describe our target encoder and how to train it along with a translation model.
The target encoder learns to map target sentences to discrete domains, and we show how to use these domains to efficiently sample diverse translations.

\subsection{Target encoding}

A Neural Machine Translation (NMT) model is composed of a source encoder $\srcenc$, and a decoder $D$.
Given a dataset $\mathcal{D}$ of pairs $(x, y)$ of source sentences and their target translations,
a standard encoder-decoder model is trained to minimize:

$$\E_{(x, y) \in \mathcal{D}} \Big( -\log p_{\dec}\big(y~|~\srcenc(x)\big) \Big)$$

\noindent where $p_{\dec}\big(y | \srcenc(x)\big)$ represents the probability given by the decoder $\dec$ to a target sentence $y$ to be the translation of a source sentence $x$.
In our case, we consider that we also have a target encoder $\tgtenc$, and we feed the decoder not only with an encoding of the source sentence, but also with an encoding of the target sentence. As a result, the model is trained to minimize:

$$\E_{(x, y) \in \mathcal{D}} \Big( -\log p_{\dec}\big(y~|~\srcenc(x); \tgtenc(y)\big) \Big)$$

\noindent Without architectural constraint, the decoder $\dec$ could trivially learn the identity mapping between the encoding of the target sentence $\tgtenc(y)$ and the sentence to generate $y$.
Instead, we propose to use a key-value structure for this embedding where the target encoder provides a probability for a key, or domain, and we feed the associated value to the decoder of the machine translation system.
In practice, we constraint the output of the target encoder to represent the domain probability distribution of the target sentence.
The output of the target encoder is thus a $\ndomain$-dimensional vector of probabilities $p = \tgtenc(y)$.
Since the output of the target encoder is not directly fed to the decoder, we bound the amount of information provided by the target encoder, preventing the model from learning a trivial mapping.
At test time, we cannot estimate $\tgtenc(y)$ since the target sentence $y$ is not available. Instead, we feed the decoder $\dec$ with any one-hot vector of $\mathbb{R}^\ndomain$ to generate $\ndomain$ different translations. An illustration of our model is provided in Figure~\ref{figure:model}.

\subsection{Implementation}

Our NMT model is the transformer network of~\citet{vaswani2017attention} with a dimension $d$, with a transformer encoder $\srcenc$ and a transformer decoder $\dec$. The target encoder $\tgtenc$ that we introduce in this paper is composed of a transformer encoder with the same architecture as the source encoder $\srcenc$ and other components detailed bellow.
We refer the reader to~\citet{vaswani2017attention} for the details of the architecture and describe below the specificites of our target encoder $\tgtenc$.

The output $\tgtenc(y)$ of the target encoder is a probability vector of size $\ndomain$. To obtain these probabilities, we encode the target through a transformer encoder. We take the first hidden state $h \in \mathbb{R}^d$ of the last layer of the target encoder, corresponding to the start token. We linearly map $h$ to a score vector of dimension $\ndomain$. Finally, we apply a $\softmax$ operator to obtain a vector of domain probabilities:
$$p = \tgtenc(y) = \softmax(M h)$$

In that setting, the decoder is trained with arbitrary probability vectors, which becomes problematic at test time when $p$ is set to a one-hot embedding on which the decoder may never have been trained. To prevent this train-test discrepancy, we apply a temperature on the domain scores $s$ that decreases linearly from 1 to 0 over training. When the temperature reaches 0, we have $p = \mathcal{I}(\argmax(s))$\footnote{By $\mathcal{I}(j) = (0, \dots, 0, 1, 0, \dots, 0)$, we denote the one-hot vector with 1 for $j$-th coordinate and 0 elsewhere).} (i.e. the domain with the highest score has probability $1$, the others have probability $0$) and the target encoder remains frozen during the remaining training time.

Moreover, at each training step, we randomly replace the $\softmax$ by an $\argmax$ operator with a probability $p_{\mathrm{hard}}$. In practice, we set $p_{\mathrm{hard}}~=~0.25$, which means that 75\% of the time the target encoder is trained along with the source encoder and decoder, and 25\% of the time the target encoder is only used to predict the domain with the highest probability. Overall, we have:

\begin{equation*}
  \tgtenc(y) = \begin{cases}
    \mathcal{I}(\argmax(s)),                & \mathrm{if~} 0 \leq X \leq p_{\mathrm{hard}} \\
    \softmax\big(\frac{s}{T}\big), & \textrm{otherwise}
  \end{cases}
\end{equation*}

\noindent where $X$ is a random variable from a uniform distribution, i.e., $X\sim \mathcal{U}(0,1)$.

\paragraph{Optimization.} When $T > 0$, the model is fully differentiable and the target encoder can be trained in an end-to-end fashion with the rest of the model. We found that it is also possible to use discrete operators like the Gumbel-Softmax~\citep{jang2016categorical}. This way, $\tgtenc(y)$ is always a one-hot vector and there is no train-test discrepancy. However, learning the target encoder through a discrete encoding makes optimization more difficult, and we obtained better results with a regular $\softmax$.

\paragraph{Domain input.}
To feed the target encoder output $\tgtenc(y)$ as input to the decoder $\dec$, the decoder learns a matrix of embeddings $E~=~\big[e_0, \dots, e_{\ndomain - 1} \big]~\in~\mathbb{R}^{d \times \ndomain}$ where each $e_i$ represents a different domain. Traditionally, the first input of a decoder is an embedding that corresponds to a start symbol $\langle S \rangle$. Instead, we feed as first embedding a vector $e$, where:
$$e = E p = \sum_{i=0}^{\ndomain-1}{p_i e_i}, \quad \textrm{with} \quad p = \tgtenc(y)$$

\begin{figure}[t]
  \center
  \includegraphics[width=0.6\linewidth]{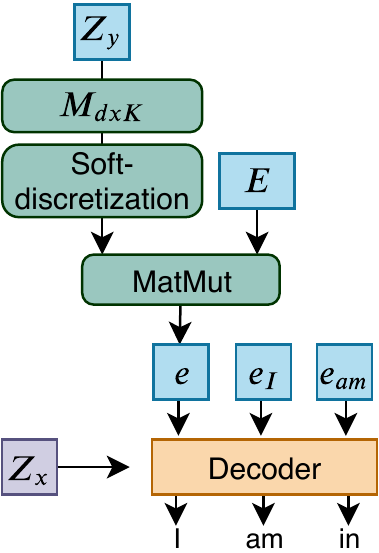}
  \caption{\textbf{Detailed illustration of our model.} $Z_y$ is the first hidden state of the output of the target transformer encoder. 
To obtain $\tgtenc(y)$, we linearly map $Z_y$ to a $\ndomain$ dimensional vector  and perform a ``soft-disctretization'' by applying either a $\softmax$ or an $\argmax$ operator. 
We then compute the target domain vector $e$ as the sum of the domain embeddings $E$ reweighted by their probabilities  contained in $\tgtenc(y)$.
The vector $e$ is fed to the decoder as the embedding of the first token, along with the source encoding $Z_x = \srcenc(x)$.}
  \label{figure:model_details}
\end{figure}

\noindent The domain embeddings $E$ are learned during training. This process is illustrated in Figure~\ref{figure:model_details}.

\subsection{Training objective}

We denote by $\theta$ the parameters of $\srcenc$, $\tgtenc$, and $\dec$. Given a mini-batch of source and target sentences $\{(x_i, y_i)\}_{1 \leq i \leq N}$, the model is trained to minimize:

\begin{equation*}
  \mathcal{L}(\theta) = \sum_{i=1}^N - \log \Big( p_{\dec}\big(y_i | \srcenc(x_i); \tgtenc(y_i)\big) \Big)
\end{equation*}

In practice, we want the decoder to properly leverage $\tgtenc(y)$, i.e., the domain information coming from the target encoder.
Without additional constraints, nothing prevents the model from collapsing to a mode where the target encoder constantly predicts the same domain, regardless of its input.
The model is then perfectly predicting its domain, which means that it receives no gradient to escape this trivial solution.

To address this issue, we add a regularization term to the training objective, to encourage the model to make a uniform usage of available domains.
In particular, we define the entropy distribution of selected domains in the mini-batch:

\begin{equation*}
  \mathcal{L}_{XE}(\theta) = - \widetilde{p} \log(\widetilde{p}), \quad \textrm{with} \quad \widetilde{p}= \frac{1}{N}\sum_{i=1}^N p_i 
\end{equation*}

\noindent where $p_i = \tgtenc(y_i)$ is the probability distribution of domains for the target sentence $y_i$. Finally, the model is trained to minimize $\mathcal{L}(\theta) - \lambda \mathcal{L}_{XE}(\theta)$, where $\lambda$ is a hyper-parameter.


\subsection{Inference}

At inference, we generate one hypothesis per domain, i.e. $\ndomain$ hypotheses. To generate the $k^{\mathrm{th}}$ hypothesis, we perform decoding by feeding $e_k$ as embedding of the start symbol. We generate translations with greedy decoding, except in Figure~\ref{figure:beamimpact}, where we combine our model with beam search decoding which leads to a different quality vs. diversity trade-off.

\section{Experiments}

In this section, we describe an evaluation protocol similar to~\citet{shen2019mixture}, and compare our approach to several baselines on $3$ MT datasets.
Then, we show the importance of different components in our model in an ablation study.

\subsection{Evaluation Metrics}

To measure both the quality and diversity of our generations, we use an evaluation protocol similar to \citet{shen2019mixture}.
The test set has multiple human reference translations which allows to measure diversity.
Formally, we denote by~$\{s_i, [r_i^1,...,r_i^P]\}_{1 \leq i \leq N}$ a multi-reference dataset, where each source sentence $s_i$ is provided with $P$ reference translations~$[r_i^1,...,r_i^P]$, and by~$[h_i^1,...,h_i^\ndomain]$ the $\ndomain$ hypotheses generated by our model for the source sentence $s_i$.

We denote by $\BLEU \big( \big\{h_i,[r_i^1,...,r_i^P] \big\}_{1 \leq i \leq N} \big)$ the corpus-level BLEU score, with $P \geq 1$ references for each hypothesis.
To measure the quality of our generations, we define:
$$\mbleu = \BLEU \bigg(   \big\{h_i^j, [r_i^1, \dots ,r_i^P] \big\}_{1 \leq i \leq N, j \in \ndomain} \bigg)$$

\noindent \mbleu measures the quality of translations for each source sentence, and for each domain.
A model that does not generate good translations for each domain will perform poorly.
To measure the diversity of translations, we use the \pairwise metric of \citet{shen2019mixture}, defined as:
$$\pairwise = \BLEU \bigg( \{h_{i, j}, [h_{i, k}]\}_{\substack{1 \leq i \leq N \\ (j, k) \in \ndomain^2, j \neq k}}    \bigg)$$\\
\pairwise \ computes the BLEU score between hypotheses of a same source sentence. A low~\pairwise ensures diversity in translations, while a~\pairwise of $100$ means that for a given source sentence, the decoder will always generate the same translation.
Overall, we want the model to have a low \pairwise while preserving a high~\mbleu score.


\subsection{Dataset}

We train and test our model on three different datasets, following \citet{shen2019mixture}. Each dataset comes with a test set with multiple human reference translations.

\paragraph{WMT'17 English-German (En-De).} We follow the same pre-processing protocol as \citet{shen2019mixture}, where we filter all training sentences with more than 80 source or target words, which results in 4.5M sentence pairs. We apply the Moses tokenizer \cite{koehn2007moses} and learn a joint BPE vocabulary with 32k codes \cite{koehn2007moses}. We take newstest2013 as a validation set, and test on a subset of 500 sentences of newstest2014 with 10 reference translations.

\paragraph{WMT'14 English-French (En-Fr).} We follow the setup of \citet{gehring2017convolutional}, which results in 36M training sentence pairs. We use a joint vocabulary of 40k BPE codes. We use newstest2012 and newstest2013 as a validation set, and test on a subset of 500 sentences from newstest2014 with 10 reference translations.

\paragraph{WMT'17 Chinese-English (Zh-En).} We follow the pre-processing setup of \citet{hassan2018achieving}. The training set is composed of 20M sentence pairs, with 48k and 32k source and target BPE vocabularies respectively. We develop on devtest2017 and evaluate on a subset of 2000 sentences of newstest2017 that comes with 3 reference translations.

\subsection{Experimental details}

In all our experiments, we consider transformers with $6$ layers, $8$ attention heads, and we set the model dimension to $d=512$. We optimize our model with the Adam optimizer \cite{kingma2014adam} with $\beta_1 = 0.9, \beta_2=0.98$ and a learning rate of $3 \times 10^{-4}$. We use the same learning rate schedule as \citet{vaswani2017attention}. We use a dropout \cite{srivastava2014dropout} of $0.1$ in the source encoder and the decoder. Following \citet{shen2019mixture}, we do not use any dropout in the target encoder. With stochasticity in the target encoder, a same target sentence tends to be mapped to different domains at different iterations, which prevents the decoder from learning the specificity of each domain, and results in identical generations with no diversity.

We use 128 GPUs for the En-Fr experiments, and 16 GPUs for the En-De and Zh-En experiments. For the En-Fr experiments, we train with mini-batches of around 450k tokens, and 55k tokens for En-De and Zh-En. We use float16 operations to speed up training and to reduce the memory usage of our models. We implement our model within the fairseq framework of \citet{ott2019fairseq}.

\begin{table*}[t]
\centering
\begin{tabular}{l c ccc  c ccc}
\toprule
    &~& \multicolumn{3}{c}{\mbleu} &~& \multicolumn{3}{c}{\pairwise}\\
    \cmidrule{3-5}\cmidrule{7-9}
    && En-De & En-Fr & Zh-En      && En-De & En-Fr & Zh-En  \\
    \midrule
    Sampling                   && \bleu{28.2} & \bleu{43.6} & \bleu{19.1} && \bleu{11.8} & \bleu{21.0} & \bleu{12.0} \\
    Beam                && \bleu{66.3} & \bleu{79.3} & \bleu{32.2} && \bleu{74.0} & \bleu{77.7} & \bleu{83.8} \\
    Diverse Beam \cite{vijayakumar2018diverse} && \bleu{60.0} & \bleu{72.5} & \bleu{31.6} && \bleu{ 53.7} & \bleu{64.9} & \bleu{66.5} \\
    MoE \cite{shen2019mixture} && \bleu{59.8}  & \bleu{72.6} & \bleu{35.7} && \bleu{48.8}  & \bleu{64.4} & \bleu{47.5} \\
    Our model                  && \bleu{55.4}  & \bleu{65.9} & \bleu{34.7} && \bleu{46.2}  & \bleu{57.3} & \bleu{52.5} \\
    \bottomrule
\end{tabular}
\caption{Results on three WMT datasets: En-De, En-Fr, Zh-En. We use $\ndomain$ = 10, 10, 3 domains respectively. We generate the same number of hypotheses as the number of references available in the multi-references datasets. Beam search is computed with beam size of $\ndomain$.}
\label{tab:all_language_results}
\end{table*}

\subsection{Baselines}

\paragraph{Sampling and Beam.} 
We report results with a sampling and a beam baseline, as well as the diverse beam method \cite{vijayakumar2018diverse}. We consider a standard NMT system (i.e. an encoder-decoder model, without target encoder or latent variable). At test time, for sampling we sample $\ndomain$ translations to generate $\ndomain$ hypotheses. For the beam search, we use a beam size of $\ndomain$ and return all hypotheses in the beam. 


\paragraph{Mixture of Experts.} We also compare against the state-of-the-art Mixture of Experts (MoE) model of \citet{shen2019mixture}, with online responsibility update, uniform prior, shared parameters and hard assignment (\textit{hMup} in their paper), which is their overall best setup. MoE model is composed of a source encoder $\srcenc$ and a decoder $\dec$. Like our model, the decoder learns a matrix of embeddings $E = \big[e_0, \dots, e_{\ndomain - 1} \big] \in \mathbb{R}^{d \times \ndomain}$ where each $e_i$ represents a different domain which is fed as first input of the decoder. Unlike us, they do not use a separate target encoder to select the domain, but consider an EM algorithm where the selected domain is the one that minimizes the reconstruction loss of the target sentence. In particular, for a mini-batch of $N$ source and target sentences $\{(x_i, y_i)\}_{1 \leq i \leq N}$, the E-step computes:
$$d_i^* = \argmax_{d \in [1, \ndomain]} p_{\dec}\big(y_i | \srcenc(x_i); d\big)$$
Then, the M-step minimizes the negative log-probability of target sentences, given their source encodings, and the selected domains:
$$\mathcal{L}(\theta) = \sum_{i=1}^N - \log \Big( p_{\dec}\big(y_i | \srcenc(x_i); d_i^*\big)\Big)$$



\medskip

We run all of these baselines with the same transformer architecture as the one used in our model. For fair comparison, we use the same optimizer, learning rate and batch size in all experiments.

\subsection{Main results}

Table~\ref{tab:all_language_results} present \mbleu and \pairwise scores for different models, on the three considered datasets. We observe that a high \mbleu score is often combined with a high \pairwise.
For instance, the beam search and sampling baselines fail at generating both diverse and high quality translations. Beam search and diverse beam search hypotheses are accurate, but lack diversity, resulting in a very similar set of hypotheses. On WMT En-De, with $\ndomain=10$, beam search gives a \mbleu score of 66.3 but a \pairwise score of 74. On the other hand, the sampling baseline generates very diverse but inaccurate hypotheses, with a \pairwise score of 11.8, but a \mbleu of 28.2.

\begin{figure}[t]
  \center
  \includegraphics[width=0.8\linewidth]{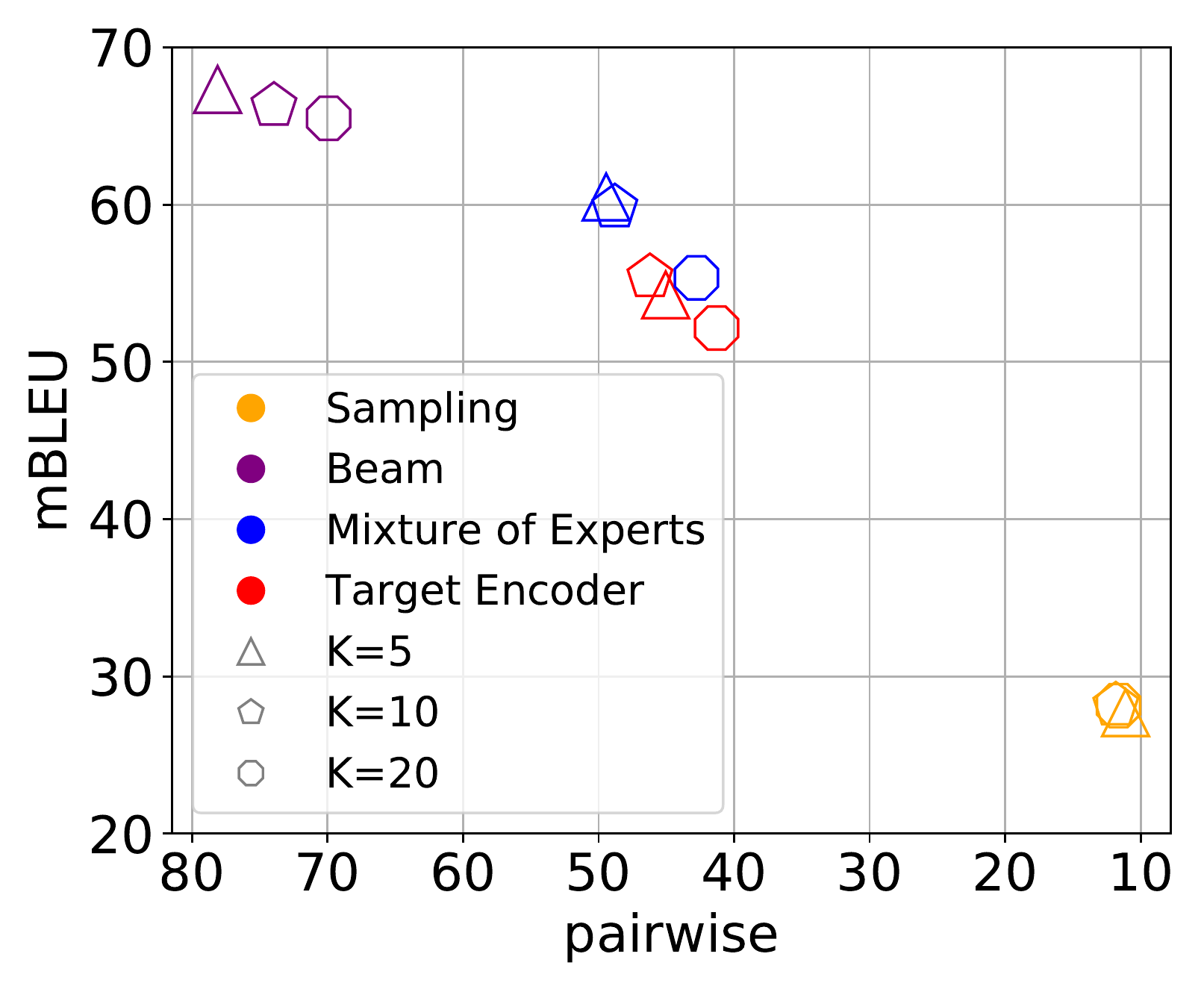}
  \caption{\textbf{Impact of the number of domains.} Results on the WMT'17 En-De dataset. We compare beam search, sampling, MoE~\cite{shen2019mixture} to our Target Encoder. In each case, we report results for $\ndomain$ = $5$, $10$ and $20$ domains. MoE and Target Encoder provide the best trade-off between quality and diversity. Compared to MoE, Target Encoder provides a lower \mbleu score, but also a lower \pairwise (i.e., more diversity).}
  \label{figure:ende_results}
\end{figure}

The Mixture of Experts and Target Encoder models have a better trade-off between diversity and quality, as shown in Figure~\ref{figure:ende_results}. Overall, our method provides more diversity than the MoE method, i.e. it obtains a lower \pairwise score, but to the detriment of a lower \mbleu score.
In Table~\ref{tab:all_language_results}, we observe that for En-De and En-Fr, our model obtains a lower \mbleu score than beam search decoding and the Mixture of Experts, but provides more diversity, with a \pairwise score of 57.3 instead of 64.4 in En-Fr. While both methods perform similarly, our approach is simpler to implement, and can easily scale to an arbitrary number of domains, as shown in the following section.

\insertgenerationscomparison

\subsection{Training speed}

\begin{figure}[t]
  \center
  \includegraphics[width=0.8\linewidth]{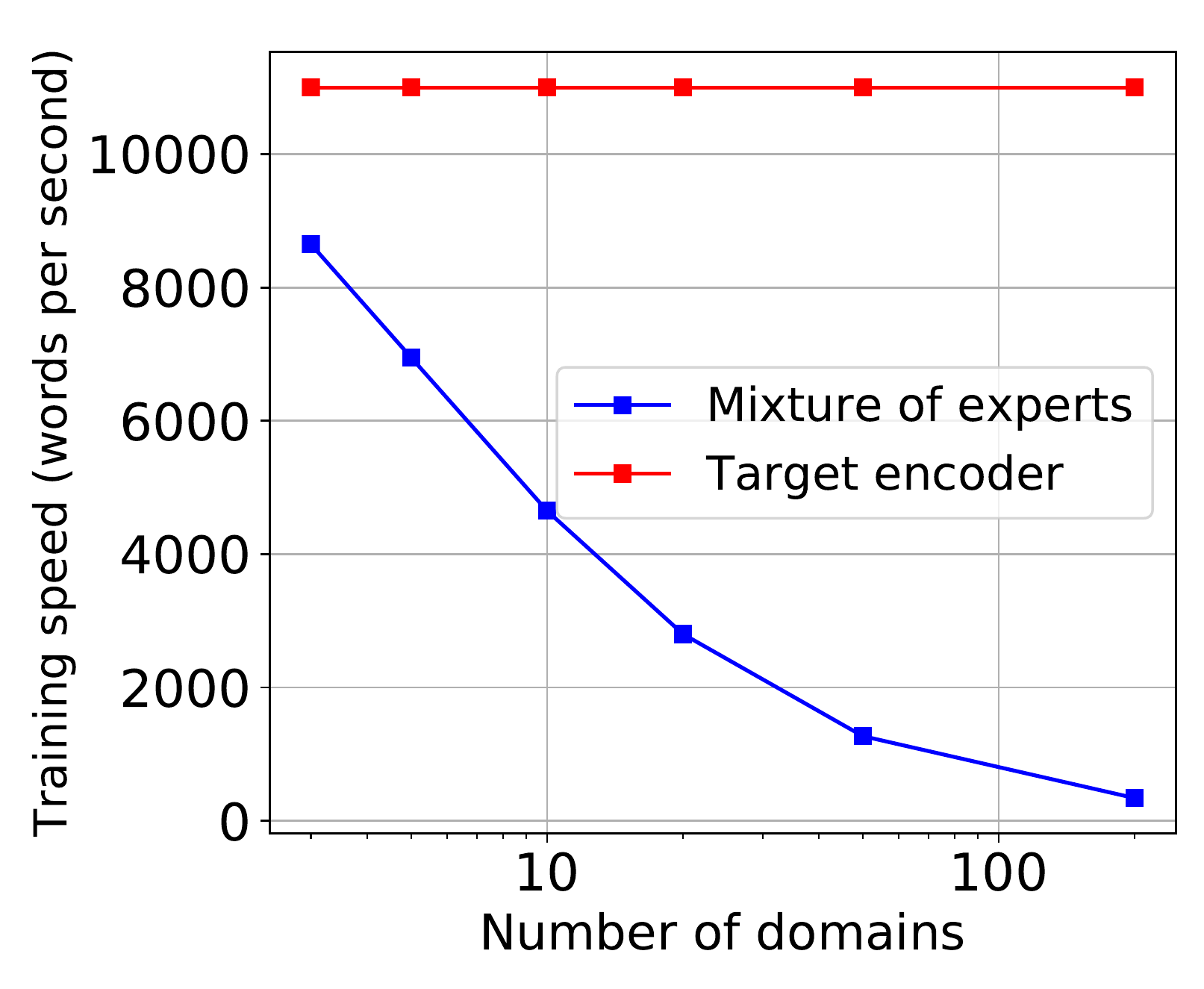}
  \caption{\textbf{Training speed.} Measured in number of words per second, for our target encoder model and the Mixture of Experts model of \citet{shen2019mixture}, for different number of domains ($\ndomain=3$, 5, 10, 20, 50, 200). The training speed of the target encoder model is constant while the Mixture of Experts model training speed decreases with the number of domains.}
  \label{figure:speed}
\end{figure}

The training speed of our method is independent of the number of domains. In contrast, the training speed of the MoE model of \citet{shen2019mixture} decreases drastically when the number of domains increases. Indeed, the MoE model requires to perform $\ndomain$ forward passes to determine the best domain. In Figure~\ref{figure:speed}, we compare the training speed of both models for $\ndomain=3$, 5, 10, 20, 50 and 200. Unlike the MoE model, using a target encoder allows generalization to an arbitrary number of domains.

\subsection{Ablation study}

\begin{figure}[t]
  \centering
  \includegraphics[width=0.8\linewidth]{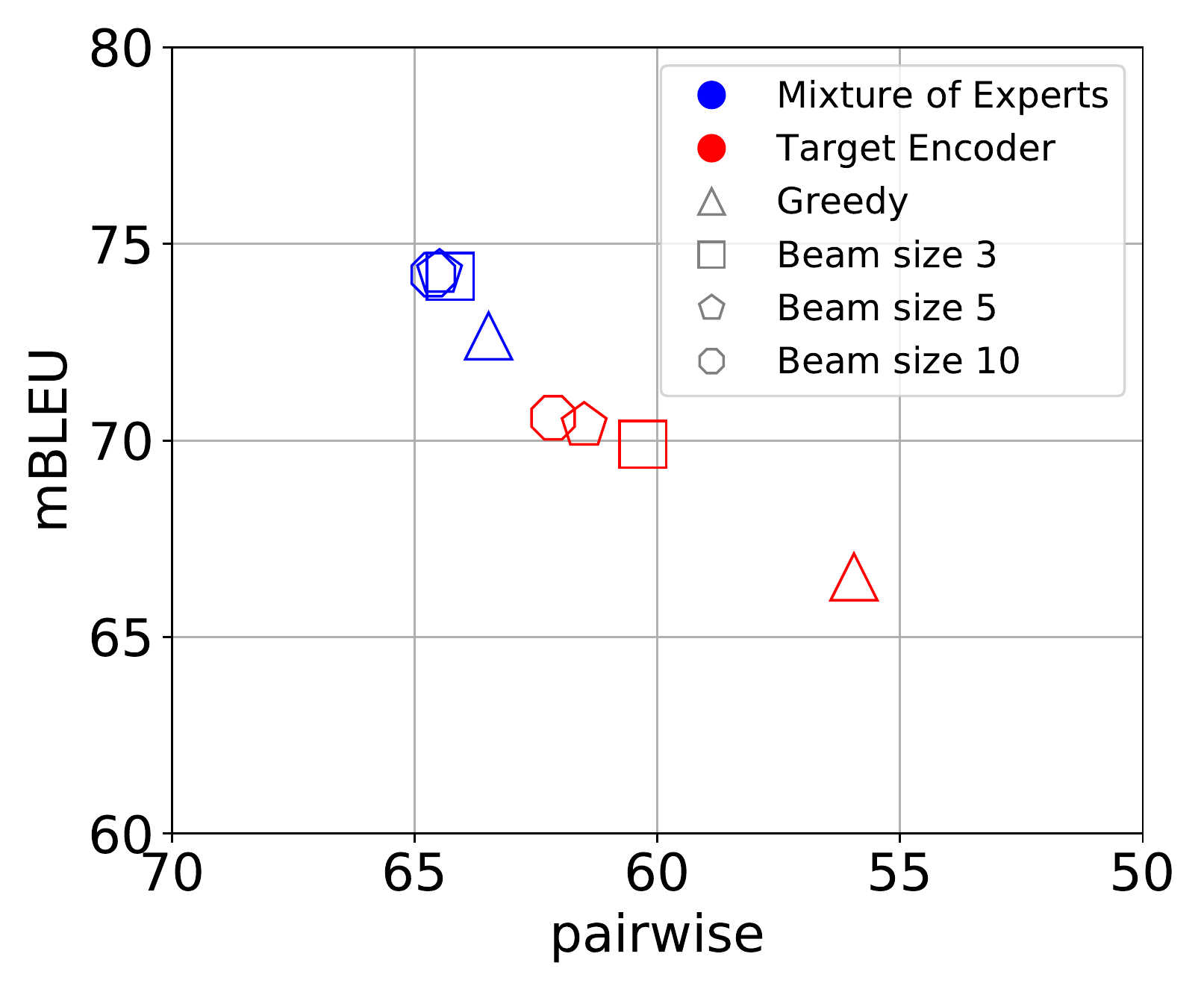}
  \caption{\textbf{Beaming search decoding.} Results on the WMT'14 En-Fr dataset for $\ndomain=10$ domains. We study the impact of decoding greedily and beam search, for beam sizes of $3$, $5$ and $10$. 
Beam search increases both the \mbleu and the \pairwise scores, i.e., it provides higher quality translations, but with lower diversity.}
  \label{figure:beamimpact}
\end{figure}

\paragraph{Beam search.} In Figure~\ref{figure:beamimpact}, we study the impact of decoding with beam search instead of greedy decoding. Using beam search improves the quality of translations, but deteriorates the diversity. Combining a target encoder model with a beam search pushes towards the same trade-off of quality-diversity as the greedy MoE model.

\insertgenerationszhen

\paragraph{Domain regularization.} Without any regularization on the domain probabilities, i.e. when $\lambda=0$, we sometimes encounter the ``collapse'' scenario where at training time all target sentences are mapped to the same domain. As a result, only the embedding associated to that domain is trained, and at test time, every sentence generated from another (and untrained) domain embedding will be invalid. This means that only one of the $\ndomain$ generated hypotheses will be valid, leading to a very poor \mbleu.
Conversely, when $\lambda$ is too high, the regularization term becomes predominant and the target encoder primarily focuses on maximizing the domain usage entropy, rather than on minimizing the decoder reconstruction loss. As a result, the target encoder uniformly maps target sentences to all available domains, but the domains do not contain any information about target sentences. This way, the decoder learns to ignore the domain, and will always output the same translation, independently of the input domain, which results in a \pairwise score close to 100 (i.e. there is no diversity). In practice, we found that setting $\lambda = 0.1$ or $\lambda = 1$ leads to similar results, and is enough to prevent the collapse scenario.

\paragraph{Source versus target encoding.}
In this experiment, we change the input of our target encoder to probe where the source of diversity in our model comes from.
In particular, it is possible that the diversity captured by our model is indirectly coming from the source sentences through the target sentences.
We test this hypothesis by replacing the input of the target encoder by the source sentence.
This model is identical to ours beside the change in the input of the target encoder.
In that setting, on WMT'17 En-De, when using 10 domains, we obtain a \mbleu score of \bleu{66.46801}, and a \pairwise BLEU of \bleu{97.19711}, which means that the model was not able to learn anything specific about each domain, and the decoder simply ignores the domain information. The fact that learning the domain from the input sentence does not work well is expected, as this information is already encoded in the source encoding $z_x$. This validates that learning the diversity form the target domains is important.
It also suggests that the diversity that our model learns is inherent to the target domain, and does not come from the source domain indirectly.
Finally, both models have the same number of parameters, suggesting that the gain in performance is not only caused by the additional parameters.

\subsection{Qualitative analysis}

Table~\ref{tab:generations_deen} provides examples of generations by our model on the WMT'17 Zh-En dataset. For each Chinese source sentence, we provide one English human translated reference, and translations by our model for three different domains.
We observe that the model generates high-quality translations with high diversity. Unlike beam search decoding, that tends to return similar hypotheses with only minor differences in the suffix \cite{ott2018analyzing}, our model is able to generate diverse translations with very different prefixes, even for long sentences.

\section{Conclusion}

In this paper, we presented an efficient way to sample diverse translations by adding a discrete target encoder to a NMT model.
The discrete representation allows to change the domain of the translation and can be trained without supervision.
The advantages of using a discrete encoder is that it is both general and scales with the number of domains with no additional computational time.
In the future, we plan to test our discrete target encoder to diversify generations in other domains, such as language modeling, image captioning or image inpainting.


\clearpage

\bibliography{acl2020}

\begin{thebibliography}{31}
\expandafter\ifx\csname natexlab\endcsname\relax\def\natexlab#1{#1}\fi

\bibitem[{Bengio et~al.(2015)Bengio, Vinyals, Jaitly, and
  Shazeer}]{bengio2015scheduled}
Samy Bengio, Oriol Vinyals, Navdeep Jaitly, and Noam Shazeer. 2015.
\newblock Scheduled sampling for sequence prediction with recurrent neural
  networks.
\newblock In \emph{Advances in Neural Information Processing Systems}, pages
  1171--1179.

\bibitem[{Bowman et~al.(2015)Bowman, Vilnis, Vinyals, Dai, Jozefowicz, and
  Bengio}]{bowman2015generating}
Samuel~R Bowman, Luke Vilnis, Oriol Vinyals, Andrew~M Dai, Rafal Jozefowicz,
  and Samy Bengio. 2015.
\newblock Generating sentences from a continuous space.
\newblock \emph{arXiv preprint arXiv:1511.06349}.

\bibitem[{Cao and Clark(2017)}]{cao2017latent}
Kris Cao and Stephen Clark. 2017.
\newblock Latent variable dialogue models and their diversity.
\newblock \emph{arXiv preprint arXiv:1702.05962}.

\bibitem[{Collobert et~al.(2019)Collobert, Hannun, and
  Synnaeve}]{collobert2019fully}
Ronan Collobert, Awni Hannun, and Gabriel Synnaeve. 2019.
\newblock A fully differentiable beam search decoder.
\newblock \emph{arXiv preprint arXiv:1902.06022}.

\bibitem[{Dai et~al.(2017)Dai, Fidler, Urtasun, and Lin}]{dai2017towards}
Bo~Dai, Sanja Fidler, Raquel Urtasun, and Dahua Lin. 2017.
\newblock Towards diverse and natural image descriptions via a conditional gan.
\newblock In \emph{Proceedings of the IEEE International Conference on Computer
  Vision}, pages 2970--2979.

\bibitem[{Gehring et~al.(2017)Gehring, Auli, Grangier, Yarats, and
  Dauphin}]{gehring2017convolutional}
Jonas Gehring, Michael Auli, David Grangier, Denis Yarats, and Yann~N Dauphin.
  2017.
\newblock Convolutional sequence to sequence learning.
\newblock In \emph{Proceedings of the 34th International Conference on Machine
  Learning-Volume 70}, pages 1243--1252. JMLR. org.

\bibitem[{Gu et~al.(2017)Gu, Cho, and Li}]{gu2017trainable}
Jiatao Gu, Kyunghyun Cho, and Victor~OK Li. 2017.
\newblock Trainable greedy decoding for neural machine translation.
\newblock \emph{arXiv preprint arXiv:1702.02429}.

\bibitem[{Hassan et~al.(2018)Hassan, Aue, Chen, Chowdhary, Clark, Federmann,
  Huang, Junczys-Dowmunt, Lewis, Li et~al.}]{hassan2018achieving}
Hany Hassan, Anthony Aue, Chang Chen, Vishal Chowdhary, Jonathan Clark,
  Christian Federmann, Xuedong Huang, Marcin Junczys-Dowmunt, William Lewis,
  Mu~Li, et~al. 2018.
\newblock Achieving human parity on automatic chinese to english news
  translation.
\newblock \emph{arXiv preprint arXiv:1803.05567}.

\bibitem[{Jain et~al.(2017)Jain, Zhang, and Schwing}]{jain2017creativity}
Unnat Jain, Ziyu Zhang, and Alexander~G Schwing. 2017.
\newblock Creativity: Generating diverse questions using variational
  autoencoders.
\newblock In \emph{Proceedings of the IEEE Conference on Computer Vision and
  Pattern Recognition}, pages 6485--6494.

\bibitem[{Jang et~al.(2016)Jang, Gu, and Poole}]{jang2016categorical}
Eric Jang, Shixiang Gu, and Ben Poole. 2016.
\newblock Categorical reparameterization with gumbel-softmax.
\newblock \emph{arXiv preprint arXiv:1611.01144}.

\bibitem[{Kaiser et~al.(2018)Kaiser, Roy, Vaswani, Parmar, Bengio, Uszkoreit,
  and Shazeer}]{kaiser2018fast}
{\L}ukasz Kaiser, Aurko Roy, Ashish Vaswani, Niki Parmar, Samy Bengio, Jakob
  Uszkoreit, and Noam Shazeer. 2018.
\newblock Fast decoding in sequence models using discrete latent variables.
\newblock \emph{arXiv preprint arXiv:1803.03382}.

\bibitem[{Kingma and Ba(2014)}]{kingma2014adam}
Diederik Kingma and Jimmy Ba. 2014.
\newblock Adam: A method for stochastic optimization.
\newblock \emph{arXiv preprint arXiv:1412.6980}.

\bibitem[{Kingma and Welling(2013)}]{kingma2013auto}
Diederik~P Kingma and Max Welling. 2013.
\newblock Auto-encoding variational bayes.
\newblock \emph{arXiv preprint arXiv:1312.6114}.

\bibitem[{Koehn et~al.(2007)Koehn, Hoang, Birch, Callison-Burch, Federico,
  Bertoldi, Cowan, Shen, Moran, Zens et~al.}]{koehn2007moses}
Philipp Koehn, Hieu Hoang, Alexandra Birch, Chris Callison-Burch, Marcello
  Federico, Nicola Bertoldi, Brooke Cowan, Wade Shen, Christine Moran, Richard
  Zens, et~al. 2007.
\newblock Moses: Open source toolkit for statistical machine translation.

\bibitem[{Kumar and Byrne(2004)}]{kumar2004minimum}
Shankar Kumar and William Byrne. 2004.
\newblock Minimum bayes-risk decoding for statistical machine translation.
\newblock Technical report.

\bibitem[{Li et~al.(2016)Li, Monroe, and Jurafsky}]{li2016simple}
Jiwei Li, Will Monroe, and Dan Jurafsky. 2016.
\newblock A simple, fast diverse decoding algorithm for neural generation.
\newblock \emph{arXiv preprint arXiv:1611.08562}.

\bibitem[{van~den Oord et~al.(2017)van~den Oord, Vinyals
  et~al.}]{van2017neural}
Aaron van~den Oord, Oriol Vinyals, et~al. 2017.
\newblock Neural discrete representation learning.
\newblock In \emph{Advances in Neural Information Processing Systems}, pages
  6306--6315.

\bibitem[{Ott et~al.(2018)Ott, Auli, Grangier, and Ranzato}]{ott2018analyzing}
Myle Ott, Michael Auli, David Grangier, and Marc'Aurelio Ranzato. 2018.
\newblock Analyzing uncertainty in neural machine translation.
\newblock \emph{arXiv preprint arXiv:1803.00047}.

\bibitem[{Ott et~al.(2019)Ott, Edunov, Baevski, Fan, Gross, Ng, Grangier, and
  Auli}]{ott2019fairseq}
Myle Ott, Sergey Edunov, Alexei Baevski, Angela Fan, Sam Gross, Nathan Ng,
  David Grangier, and Michael Auli. 2019.
\newblock fairseq: A fast, extensible toolkit for sequence modeling.
\newblock In \emph{Proceedings of NAACL-HLT 2019: Demonstrations}.

\bibitem[{Ranzato et~al.(2015)Ranzato, Chopra, Auli, and
  Zaremba}]{ranzato2015sequence}
Marc'Aurelio Ranzato, Sumit Chopra, Michael Auli, and Wojciech Zaremba. 2015.
\newblock Sequence level training with recurrent neural networks.
\newblock \emph{arXiv preprint arXiv:1511.06732}.

\bibitem[{Serban et~al.(2017)Serban, Sordoni, Lowe, Charlin, Pineau, Courville,
  and Bengio}]{serban2017hierarchical}
Iulian~Vlad Serban, Alessandro Sordoni, Ryan Lowe, Laurent Charlin, Joelle
  Pineau, Aaron Courville, and Yoshua Bengio. 2017.
\newblock A hierarchical latent variable encoder-decoder model for generating
  dialogues.
\newblock In \emph{Thirty-First AAAI Conference on Artificial Intelligence}.

\bibitem[{Shen et~al.(2019)Shen, Ott, Auli, and Ranzato}]{shen2019mixture}
Tianxiao Shen, Myle Ott, Michael Auli, and Marc'Aurelio Ranzato. 2019.
\newblock Mixture models for diverse machine translation: Tricks of the trade.
\newblock \emph{arXiv preprint arXiv:1902.07816}.

\bibitem[{Shu et~al.(2019)Shu, Nakayama, and Cho}]{shu2019generating}
Raphael Shu, Hideki Nakayama, and Kyunghyun Cho. 2019.
\newblock Generating diverse translations with sentence codes.
\newblock In \emph{Proceedings of the 57th Annual Meeting of the Association
  for Computational Linguistics}, pages 1823--1827.

\bibitem[{Srivastava et~al.(2014)Srivastava, Hinton, Krizhevsky, Sutskever, and
  Salakhutdinov}]{srivastava2014dropout}
Nitish Srivastava, Geoffrey Hinton, Alex Krizhevsky, Ilya Sutskever, and Ruslan
  Salakhutdinov. 2014.
\newblock Dropout: a simple way to prevent neural networks from overfitting.
\newblock \emph{The journal of machine learning research}, 15(1):1929--1958.

\bibitem[{Vaswani et~al.(2017)Vaswani, Shazeer, Parmar, Uszkoreit, Jones,
  Gomez, Kaiser, and Polosukhin}]{vaswani2017attention}
Ashish Vaswani, Noam Shazeer, Niki Parmar, Jakob Uszkoreit, Llion Jones,
  Aidan~N Gomez, {\L}ukasz Kaiser, and Illia Polosukhin. 2017.
\newblock Attention is all you need.
\newblock In \emph{Advances in neural information processing systems}, pages
  5998--6008.

\bibitem[{Vijayakumar et~al.(2018)Vijayakumar, Cogswell, Selvaraju, Sun, Lee,
  Crandall, and Batra}]{vijayakumar2018diverse}
Ashwin~K Vijayakumar, Michael Cogswell, Ramprasaath~R Selvaraju, Qing Sun,
  Stefan Lee, David Crandall, and Dhruv Batra. 2018.
\newblock Diverse beam search for improved description of complex scenes.
\newblock In \emph{Thirty-Second AAAI Conference on Artificial Intelligence}.

\bibitem[{Wang et~al.(2017)Wang, Schwing, and Lazebnik}]{wang2017diverse}
Liwei Wang, Alexander Schwing, and Svetlana Lazebnik. 2017.
\newblock Diverse and accurate image description using a variational
  auto-encoder with an additive gaussian encoding space.
\newblock In \emph{Advances in Neural Information Processing Systems}, pages
  5756--5766.

\bibitem[{Wen et~al.(2017)Wen, Miao, Blunsom, and Young}]{wen2017latent}
Tsung-Hsien Wen, Yishu Miao, Phil Blunsom, and Steve Young. 2017.
\newblock Latent intention dialogue models.
\newblock In \emph{Proceedings of the 34th International Conference on Machine
  Learning-Volume 70}, pages 3732--3741. JMLR. org.

\bibitem[{Wiseman and Rush(2016)}]{wiseman2016sequence}
Sam Wiseman and Alexander~M Rush. 2016.
\newblock Sequence-to-sequence learning as beam-search optimization.
\newblock \emph{arXiv preprint arXiv:1606.02960}.

\bibitem[{Xu et~al.(2018)Xu, Zhang, Qu, Xie, and Nock}]{xu2018d}
Qiongkai Xu, Juyan Zhang, Lizhen Qu, Lexing Xie, and Richard Nock. 2018.
\newblock D-page: Diverse paraphrase generation.
\newblock \emph{arXiv preprint arXiv:1808.04364}.

\bibitem[{Zhang et~al.(2016)Zhang, Xiong, Duan, Zhang
  et~al.}]{zhang2016variational}
Biao Zhang, Deyi Xiong, Hong Duan, Min Zhang, et~al. 2016.
\newblock Variational neural machine translation.
\newblock In \emph{Proceedings of the 2016 Conference on Empirical Methods in
  Natural Language Processing}, pages 521--530.

\end{thebibliography}
\bibliographystyle{acl_natbib}

\end{document}